# Multimodal Urban Areas of Interest Generation via Remote Sensing Imagery and Geographical Prior

Chuanji Shi
Ant Group
Hangzhou, China
chuanji.scj@antgroup.com

Yingying Zhang
Ant Group
Hangzhou, China
qichu.zyy@antgroup.com

Jiaotuan Wang
Ant Group
Hangzhou, China
yunting.wjt@antgroup.com

Xin Guo
Ant Group
Hangzhou, China
bangzhu.gx@antgroup.com

Qiqi Zhu*
School of Geography and Information Engineering
China University of Geosciences
Wuhan, China
zhuqq@cug.edu.cn

## ABSTRACT

Urban area-of-interest (AOI) refers to an integrated urban functional zone with defined polygonal boundaries. The rapid development of urban commerce has led to increasing demands for highly accurate and timely AOI data. However, existing research primarily focuses on coarse-grained functional zones for urban planning or regional economic analysis, and often neglects AOI's expiration in the real world. They fail to fulfill the precision demands of Mobile Internet Online-to-Offline (O2O) businesses. These businesses require accuracy down to a specific community, school, or hospital. In this paper, we propose a comprehensive end-to-end multimodal deep learning framework designed for simultaneously detecting accurate AOI boundaries and validating AOI's reliability by leveraging remote sensing imagery coupled with geographical prior, titled AOITR. Unlike conventional AOI generation methods, such as the Road-cut method that segments road networks at various levels, our approach diverges from semantic segmentation algorithms that depend on pixel-level classification. Instead, our AOITR begins by selecting a point-of-interest (POI) of specific category, and uses it to retrieve corresponding remote sensing imagery and geographical prior such as entrance POIs and road nodes. This information helps to build a multimodal detection model based on transformer encoder-decoder architecture to regress the AOI polygon. Additionally, we utilize the dynamic features from human mobility, nearby POIs, and logistics addresses for AOI reliability evaluation via a cascaded network module. The experimental results reveal that our algorithm achieves a significant improvement on Intersection over Union (IoU) metric, surpassing previous methods by a large margin. Furthermore, the AOIs produced by AOITR have substantially enriched our AOI application library and have been successfully applied on over 10 different O2O scenarios including Alipay's face scan payment service.

## 1 INTRODUCTION

Urban area-of-interest (AOI) refers to specific functional zones of basic units [33], and each AOI is spatially aggregated by diverse geographic objects [34]. Not only are they significant in city studies, but they are also widely utilized in precision marketing and meticulous management in real-world online-to-offline(O2O) businesses. For instance, grouping orders based on AOI (such as a school, residential community, or hospital) can significantly reduce costs in food and express delivery [32]. Moreover, just like Point of interest (POI) is used for online taxi-hailing, AOI is also indispensable in various location-based services, including coupon promotions in shopping malls and identification of student users. However, the rapid development of urban areas has led to a significant shortage of AOI data in many categories, which is further exacerbated by slow updates due to heavy reliance on manual annotation [35]. The issue affects not only proprietary mapping services such as Geode and Google Map but also extends to open-source alternatives like OpenStreetMap (OSM), among others.

The generation of AOI was initially accomplished through manual drawing or traditional GIS multi-step approaches, which were both expensive and slow to update [32]. In recent years, machine learning and deep learning algorithms have been developed to mine AOI, greatly improving efficiency. Currently, there are mainly two kinds of approaches in the research on AOI detection, which includes boundary generation. The first kind, which we refer as the Road-cut algorithm, utilizes the road network to divide the region into grids with certain granularity as the boundary of AOI, and then retrieves the POIs and other information via AOI to predict the corresponding category with topic models [9, 32]. Figure 1(a) demonstrates the effectiveness of the Road-cut algorithm in detecting AOIs in the well-developed area, such as city center, as depicted by the yellow lines. However, within more complex urban regions where the road network fails to distinguish a single physical entity, the detected AOI polygon is often excessively large, as showed in Figure 1(b). Additionally, in Figure 1(c), for AOIs lacking a closed-loop complete road network in suburban areas, the AOI boundary polygon cannot be extracted through the road network. The second kind is to mine and classify urban AOI using image semantic segmentation algorithms, such as U-Net [23], DeepLab [5], GCN [13] and DETR [3]. These algorithms directly predict the label of each pixel in the remote sensing image, enabling the generation of AOI boundaries and categories simultaneously. The image semantic segmentation results demonstrate good prediction accuracy for AOIs with a single building [2, 11, 25]. However, for complicated AOIs containing various geographical elements, such as hospitals with main buildings, green belts, and parking areas, the AOI polygon detection results are subpar, and the predicted boundaries are incoherent, as shown in the shaded area in Figure 1(d-f).

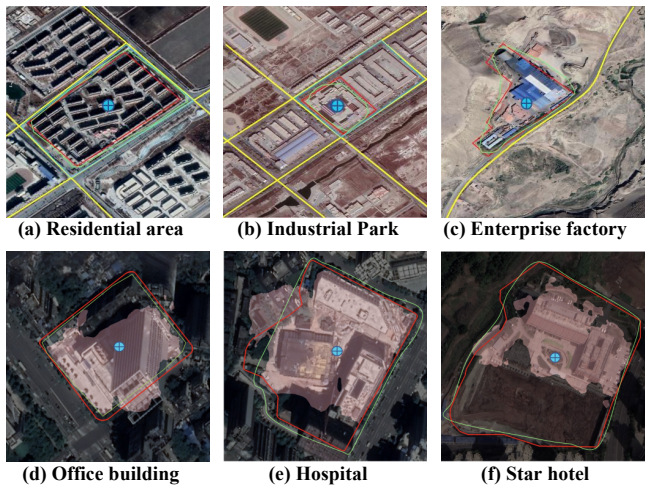

Figure 1: The predicted AOI boundaries of Road-cut algorithm and image semantic segmentation algorithm. The yellow lines represent the road network. The green lines represent the predicted AOI boundaries, and red lines represent the ground-truth boundaries.

In this paper, we propose a multimodal deep learning model to predict AOI boundaries of a pre-selected POI of specific category (called core POI), and evaluate its reliability. Our model is based on the transformer end-to-end target detection architecture, our approach involves the construction of a specialized regression head that accurately predicts the fence polygon of AOI. Additionally, we have redesigned the content query component of the transformer decoder to enhance the extraction and fusion of remote images and geographical prior. To evaluate its timeliness or reliability, we have incorporated dynamic human mobility and logistics address information into a cascaded feedforward network.

Finally, we evaluated the effectiveness of our method using the latest AOI dataset with corresponding POI, Location Based Service (LBS) datasets from our company in 2024, which contained 296854 samples across 20 categories, and the corresponding remote sensing images from Google Earth. In comparison to road network segmentation and semantic segmentation models, our model achieves a significantly higher detection accuracy with an IoU index of 0.726. It significantly surpasses the performance of the aforementioned two methods, which only achieve a maximum IoU level of 0.532. It also performs better than a recent study on AOI generation based on a similar transformer-based architecture, which reported a best IoU level of 0.622 [41]. The main contributions of this article:

- We introduced a specialized node sampling method of AOI polygon centered on the core POI. This method provides better detection results and smoother boundaries, compared to image identification or segmentation models based on pixel classification.
- Our proposed approach is an end-to-end multimodal algorithm for AOI detection and expiration identification. To the best of our knowledge, we are the first to develop a method that directly predicts the polygonal boundaries of the core POI, along with geographical prior.
- The AOITR's performance was assessed using a dataset of nearly 300,000 samples across 20 distinct categories from Gaode Map and OpenStreetMap. It achieved the highest IoU metric to date, outperforming previous methods including the Road-cut algorithm, various image segmentation algorithms, and other transformer-based methods. Additionally, the AOIs generated by AOITR have significantly enriched our AOI dataset and have been effectively applied in multiple O2O business scenarios.

## 2 RELATED WORK

Urban functional zones, also known as AOI, or ROI (region of interest), have been studied in the past mainly for urban planning, historical change, regional economy or traffic analysis. The research methods used were primarily based on traditional GIS analysis. For example, Siliang Chen [6] researched the historical change of city land use primarily using GIS methods, while [7, 12, 18] provide a good review of related work based on clustering algorithms. In other studies, community detection methods [21], such as non-negative matrix factorization (NMF), were widely adopted to describe urban functional regions based on interaction intensities. Early research by [19, 28, 36] utilized the Morphology method to mine functional areas through remote sensing data.

In recent years, with the development of urban computing, research on detecting AOIs can be classified into two approaches. Yu Zheng et al. [31, 32] utilized the road network to divide the region into grids with appropriate granularity through morphological means. The inner POIs and human mobility were then recalled, and the grid theme was extracted via topic models, such as LDA and TF-IDF [9], to generate the final AOI. Alternatively, [10, 29] used a graph convolutional neural network to classify the topic of each grid segmented by the road network, while [16, 20, 27, 30] used the standard grid to replace the grid cut by the road network and then combined distribution characteristics of nearby POIs and taxi origin/destination (O/D) flows or User Check-in data to identify specific types of functional regions.

Another approach to mining and classifying urban AOI is through remote sensing data, or by combining remote sensing images with other multimodal data. [24] proposed a graph-based algorithm that incorporates multisource data to classify AOI. Similarly, [14, 22] respectively employed a tree model and support vector machine, integrating remote sensing data, trajectory data, and social media data to classify AOI in multiple steps. Zhu et al. [38, 39, 40] utilized the bag-of-visual-words (BOVW) model and multi-feature fusion probabilistic topic model for scene classification of high spatial resolution remote sensing imagery. [4] proposed a joint multi-image saliency (JMS) algorithm to extract ROIs in a set of optical multispectral remote sensing images. Other algorithms that classify urban areas based on remote sensing data are comprehensively compared in [26]. Zhao et al. [37] introduced how to use OSM's open-source road network and label data for scene recognition research. However, our method departs from these existing approaches as we do not rely on a richer multi-semantic dataset to detect AOIs. Instead, we design an end-to-end multimodal AOI polygonal boundaries detecting model to predict the accurate polygon of AOI based on the core POI. This is different from a larger region containing multiple AOI entities cut by the road network (mainly high-level roads), or sense classification based on fixed-size remote sensing images [8, 17]. We utilized the transformer architecture to develop our approach, as it demonstrates strong performance across a range of object detection scenarios and offers seamless adaptability to other tasks, including panoramic segmentation [3]. However, despite achieving good results on distinguishable geographic atom objects [1], like buildings or school playgrounds, the transformer-based end-to-end target detection model (DETR) [3] or other state-of-the-art image segmentation algorithms, such as U-Net [23], UperNet [28] or E2EC [15], are not well-suited for AOI polygon detection. This is because different categories of AOI typically cover multiple basic geographic objects, forming a set of multiple elements with specific patterns. For example, a school includes playgrounds, teaching buildings, canteens, etc., while an industrial park includes office buildings, parking fields, or blank areas, and so on.

## 3 Methodology

Our multimodal AOI detection model is based on transformer encoder-decoder architecture, named AOITR. The generation of the AOI polygon is the core component of AOITR, as shown in the upper part of Figure 2. AOITR employs a transformer with multimodal content query in the decoder, paired with a customized prediction head module. The token projection and query projection are utilized to simultaneously capture the image and geographical features from remote sensing imagery and geographical prior respectively.

The model is further enhanced by incorporating a feedforward neural network (FFN) that utilizes dynamic human mobility trajectories, nearby POIs and extracted POIs from logistics addresses to evaluate the generated AOI's reliability, as shown in the lower part of Figure 2. The high- and low-quality AOIs are labeled as positive and negative samples respectively to train the model's capacity to discern the reliability of AOIs. This process ensures that only the predictions meeting our high-quality standards are included in our final AOI application library.

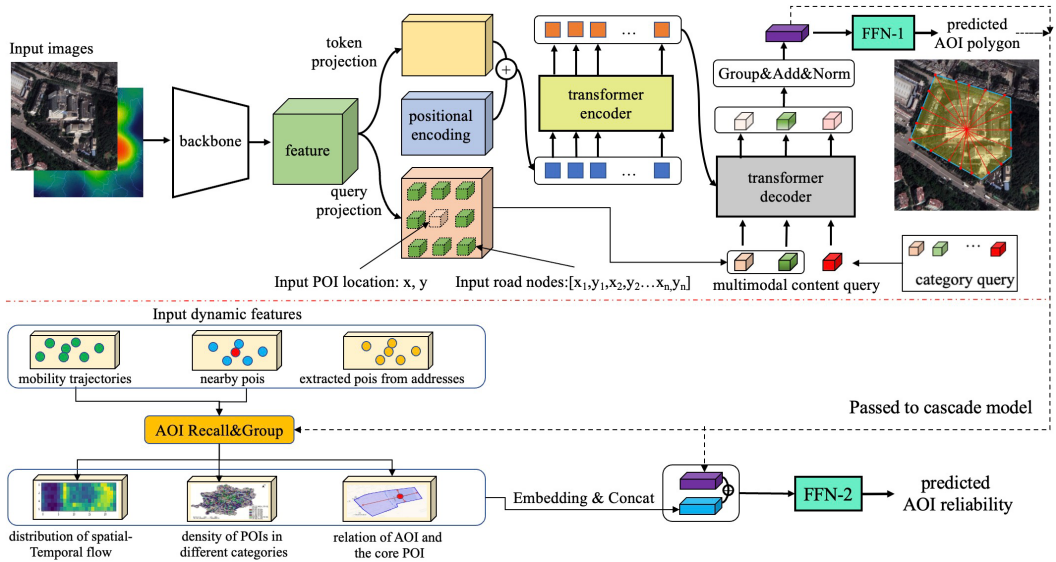

**Figure 2: Multimodal detection transformer for AOI prediction. The red dotted line above is the transformer for AOI polygon generation, and below is the module for AOI reliability evaluation.**

## 3.1 Transformer for AOI polygon generation

*3.1.1 Encoder.* The encoder follows the standard architecture of the transformer. The input to the encoder is a combination of the token projection, obtained from the image convolution layer, and the positional encodings. Each encoder layer comprises a multi-head self-attention module and an FFN.

*3.1.2 Decoder and the regression head.* The decoder adheres to the transformer's vanilla design mostly, with the only distinction in the decoder queries. There are two kinds of queries: the positional query (positional encodings) and the content one. For the prediction of the AOI polygon, the nearby road network nodes, and the associated POI distribution are crucial reference information, which we refer as geographical prior. Thus, as shown in Figure 3(b), we have designed a multimodal content query module, which consists of three semantic query parts: the location embedding $P_L \in R^{b \times 1 \times d_{model}}$ and category embedding $P_C \in R^{b \times 1 \times d_{model}}$ of the core POI $P(x,y)$, and the node locations embedding of the road network $R_L \in R^{b \times N \times d_{model}}$. The first-dimension b represents batch size, the second dimension 1 or N represents the number of reference points, and the last dimension $d_{model}$ is the embedding dimensionality.

After passing through the transformer decoder network, the three content query outputs divide to two groups: the $P_L$ & $R_L$ are added to regress the AOI polygon initial N coordinates $\hat{p}_{init}$, while $R_c$ is used to regress the offset values of each point $\hat{p}_{res}$. Finally, they are added to obtain the final outputs of the regression head, representing the N coordinates $\hat{p}$ of the AOI polygon.

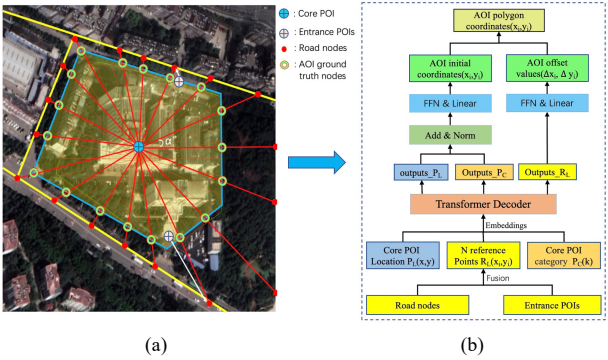

**Figure 3: The node sampling method of AOI polygons, the content query of transformer decoder and the head for AOI polygon generation. (a) The distribution of sampled AOI polygonal boundary points and geographical prior reference points. (b) The method for predicting AOI polygon coordinates using multimodal content query.**

*3.1.3 The input image and multimodal content query.* Figure 3(a) shows the input image and how we sample geographic prior information, including the core POI, road nodes, and other associated POIs. During model training, the initial remote sensing images is 256×256 pixels, we fuse and stitch 10×10 adjacent images around the core POI to create the input image of the model. The actual geographic coverage is approximately 800m×600m, thereby ensuring that the model can cater to the prediction requirements of various categories of AOI.

To represent one AOI polygon, we sample N fixed nodes from the intersections of the equiangular ray starting from the core POI location and the AOI polygon boundaries. Compared to the equidistance sampling method of AOI polygon nodes, the equiangular sampling method can better ensure the balance of the boundary points. The former is likely to cause too few points to be sampled on the shorter side. Similarly, the nodes of road network are sampled in the same way. We also retrieve the entrance POIs belonging to the core POI as reference points, which can help to the identify the actual range of AOI boundary, such as gates of a community or a school. The number of entrance POIs is M (M≤N) of non-equal length. To facilitate model processing, the M reference points will replace the nearest M associated road reference points. Meanwhile, reference points beyond the boundaries of remote sensing images will be truncated. In the end, the total number of reference points remains N. During the prediction phase, the model maintains consistency with the input used during training, with the exception of the N boundary coordinates of the AOI. These N boundary coordinates of the AOI are predicted and output by the model.

The lower part of Figure 3 (b) illustrates the process of constructing content query through embeddings, utilizing the geographic prior information. The POI category query $P_C$ is a learnable embedding, while the location queries $P_L$ and $R_L$ are sampled from the query projection through bilinear interpolation, which can be simply defined as:

$$f(P) = (x_2 - x)(y_2 - y)f(Q_{11}) + (x - x_1)(y_2 - y)f(Q_{21}) \\ + (x_2 - x)(y - y_1)f(Q_{12}) + (x - x_1)(y - y_1)f(Q_{22}) \quad (1)$$

where $Q_{11}$, $Q_{21}$, $Q_{12}$, $Q_{22}$ are the adjacent points around point P on the image, respectively.

*3.1.4 The regression loss.* The vectors $p \in R^{N \times 2}$ and $\hat{p} \in R^{N \times 2}$ represent the ground truth AOI polygon coordinates and predicted coordinates, respectively, where N is equal to the number of reference points. For the prediction with index $\sigma(i)$, we define $\mathcal{L}_{aoi} = \mathcal{L}_{iou}(p_{\sigma(i)}, \hat{p}_i)$. Considering the non-differentiability of the IoU between polygons. As a result, the loss is directly calculated as the L1 loss between the ground truth AOI polygon and the predicted AOI polygon, which is defined as:

$$L_1(p_{\sigma(i)}, \hat{p}_i) = \|p_{\sigma(i)} - \hat{p}_i\|_1 \quad (2)$$

## 3.2 AOI reliability evaluation Module

Due to the potential inaccuracy or expiration of POI, the selected core POI may be corrupted or outdated. Therefore, we additionally employ the dynamic spatio-temporal features around the AOI, including the spatio-temporal distribution of the user's mobile location, the keywords of recent logistics addresses, and other sub-POIs collected recently, to verify the reliability of the generating AOI. This not only ensures the authenticity of the core POI itself but also verifies the reasonableness of the predicted AOI polygon.

As depicted in the lower half of Figure 2, the cascade module is a binary classification model based on an FFN with a cross-entropy

Multimodal Urban Areas of Interest Generation via Remote Sensing Imagery and Geographical Priorloss. Constructing training dataset is a challenging task due to the lack of negative samples. The positive samples are selected from our internal AOI library, while the negative samples come from two origins: expired AOIs from our offline investigation and the predicted AOIs with an IoU less than a certain threshold, indicating significantly different predictions from ground truth. Another crucial part is how to extract relatively dynamic features in or nearby the predicted AOI polygon. This involves two parts: the embedding of image features directly passed from the transformer decoder outputs and the extracted of features from multi-semantic nearby POIs and human mobility, which contain POI features and dynamic LBS points distribution features.

In the cascade module, except for image features that directly passed from the transformer decoder, the input features contain POI features and dynamic LBS points distribution features.

(a) POI features:
- The basic features for the core POI and predicted AOI polygon include category, polygon area, and the central angle between the core POI location and AOI polygon.
- First, we select the core POI dataset C ($c_1, c_2, c_3 \ldots c_m$), and entrance POIs $S_c$. The logistics address POIs set $L_c$ is obtained from the desensitized delivery address of Alipay users for a week, standardized by the NER API provided by Alibaba Cloud (https://addrp.console.aliyun.com/overview). $L_c$ contains repeated POIs, and $distinct\,(L_c) \in S_c$, $L_c$ is used to reinforce the weight of frequently used POIs, which are considered more time-sensitive and reliable. Then, we calculate the POI distribution features within the AOI polygon $P$ ($p_1, p_2, p_3 \ldots p_m$):

$$\partial_{p_i|c_i} = \frac{(S_{c_i} + L_{c_i}) \in p_i}{S_{c_i} + L_{c_i}} \quad (3)$$

$$\beta_{p_i|c_i} = \frac{A_{convex}\,(S_{c_i}) \in p_i}{A\,(p_i)} \quad (4)$$

$$\delta_{p_i|c_i} = C \in p_i D \quad (5)$$

Here, $\partial_{p_i|c_i}$ refers to the rate at which locations that logically belong to the core POI are actually located in the predicted $p_i$. $\delta_{p_i|c_i}$ refers to the IoU between the area of the convex polygon of $S_{c_i}$ and the area of $p_i$. $\delta_{p_i|c_i}$ refers to the number of core POI located in the predicted $p_i$, which is expected to have only one core POI $c_i$.

(b) Dynamic LBS points distribution features：Different urban functional areas in a city have different human activity patterns. For example, active hours in residential areas are usually concentrated from 6:00 pm to 8:00 am, while in office areas, they are concentrated from 9:00 am to 6:00 pm. Therefore, we calculated the spatio-temporal distribution characteristics $D_{w,h}$ of the activity flow in the predicted polygon. Figure 4 below shows the flow density distribution of different functional areas at different time periods.

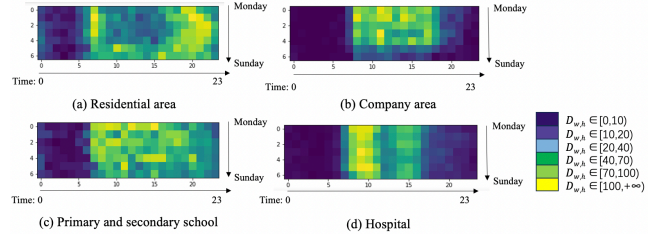

Figure 4: The flow density distribution of 5 functional areas at different time periods.

## 4 Experiments and results

### 4.1 Data Settings and Training detail

*4.1.1 Datasets*. Figure 5 illustrates an example distribution of the input multimodal data used in the AOITR and the resultant AOIs we generated in the city of Shanghai.

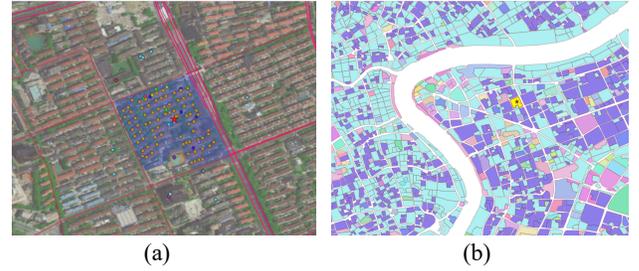

Figure 5: The raw multimodal data for AOI generation and the visualization of AOI polygons in Pudong, Shanghai.

Specifically, we used the following multimodal datasets for model training and evaluation:

(a) POI data: we selected 10 Chinese cities (Beijing, Shanghai, Shenzhen, Guangzhou, Hangzhou, Nanjing, Wuhan, Tianjin, Chengdu and Chongqing) POI datasets in the year 2024 form our company. The attributes of the POI data include name, category, longitude and latitude, and parent-child relationship, which help us identify core POIs and subsidiary POIs more accurately.

(b) AOI data: The AOI data was constructed from the core POI datasets with boundaries, which were manually drawn over the years. The detail of Merged POI/AOI categories are shown in Table 1.

(c) Remote sensing images: The remote sensing data for the same 10 cities was downloaded from Google Earth. All original images are 256×256 pixels, and the spatial resolution is about 0.3 m to 1.0 m. We stitched together 10×10 remote sensing images centered on the core POI to create the input image, and cropped the image to a latitude and longitude range of 0.006 degrees. The Figure 1 shows the complete images of different urban functional area.

(d) Road networks: We obtained the road networks of 10 Chinese cities in 2024 from OpenStreetMap and selected all major roads and external roads, excluding internal roads, resulting in a total of approximately 20 million road segments.
(e) Mobility LBS points: The dynamic points were obtained from the Alipay APP when users used the LBS we provided. There were nearly 1 billion points per day in the selected 10 cities, and we selected all April desensitization data in 2024.
(f) Logistics address: The logistics address we used come from the desensitized delivery address of Alipay users who paid for online shopping, in total about 2 million for a month in the 10 cities. We also selected the same period data as mobility LBS points.

**Table 1: POI/AOI Category taxonomy code and number of samples.**

| Code | POI category | Samples | Code | POI category | Samples |
|------|-------------|---------|------|-------------|---------|
| 0 | Exhibition hall, cultural center | 2324 | 10 | University, vocational technical school | 3055 |
| 1 | Hospital | 3529 | 11 | City square, park square | 1154 |
| 2 | Natural scenic spots, gardens | 5903 | 12 | Office building, industrial building | 29326 |
| 3 | Gas/other energy station | 7468 | 13 | Primary and secondary school | 7056 |
| 4 | Government agencies, civil society organization | 2569 | 14 | Residential area | 48103 |
| 5 | Kindergarten | 7106 | 15 | Train station, airport | 897 |
| 6 | Star hotel | 18903 | 16 | Shopping mall | 5273 |
| 7 | Leisure and entertainment place | 2835 | 17 | Factory | 3803 |
| 8 | Industrial Park | 9130 | 18 | Company | 61013 |
| 9 | Car service area, parking lot | 17379 | 19 | Other | 60028 |

*4.1.2 Experimental design.* Our AOI polygon generation model, as shown in Figure 2, consists of two parts: the main multimodal deep regression model for AOI polygon generation and a cascade module for AOI reliability evaluation. The former's input comprises the core POI's coordinates, corresponding remote sensing imagery, and a sample of N AOI polygon coordinates along with N reference points, as extracted using the method depicted in Figure 3. By default, N is set to 24 and $d_{model} = 256$. Finally, we split the above AOI samples into two parts, with an 8:2 ratio for the training and validation datasets, respectively.

The AOI-related features of the cascade module are calculated using the method described in section 3.2. As the cascade module is essentially a binary classification model, the number of positive and negative samples is the same. 100,000 positive samples are selected from our AOI library, which are frequently used. Meanwhile, half of the negative sample dataset comes from expired AOIs. For the remaining negative samples, we use predicted AOIs from the AOITR with an IoU less than 0.75.

*4.1.3 Training platform and hyperparameters.* The backbone of AOITR is set to swin-base. To train our multimodal AOI detection method with large datasets, we deployed our training on 8 Nvidia A100 GPUs. We trained the AOITR and cascade module for 50 epochs.

## 4.2 Evaluation Metrics

The mIoU (mean intersection over union) is a crucial and direct indicator used to measure the quality of predicted AOI polygon, which was defined as follows:

$$mIoU = \frac{1}{n}\sum_{i=1}^{n}\frac{A_O}{A_U} \quad (6)$$

where $A_O$ indicates the overlap area between the predicted polygon and ground truth polygon, and $A_U$ is the union area of the predicted polygon and ground truth polygon. In the evaluation module of AOITR, we use the widely adopted precision and recall metrics to evaluate the reliability of predicted AOIs.

## 4.3 Results

*4.3.1 The prediction results of different methods.* In this study, we compared our algorithms with two types of existing methods: Road-cut [31] and three widely cited image semantic segmentation algorithms, including U-Net [23], the base model DETR [3] and another semantic segmentation algorithm called UperNet [28]. These image semantic segmentation algorithms directly predict the label of each pixel in the remote sensing image to obtain the AOI mask. Then, the predicted coordinates of the AOI polygon are obtained by extracting the contour coordinates of the mask.

Table 2 shows the all results of our proposed AOITR and the four baseline methods for detecting the AOI polygon under the same setting as mentioned above. High-IoU refers to IoU>0.75, which meets the requirements of most real-world business scenarios. In Figure 6, we show the predicted mIoU for the merged 20 AOI categories.

**Table 2. The predicted mIoU of different methods.**

| Models | Input data | Total mIoU | Total high-IoU rate | Training Time (h) |
|---|---|---|---|---|
| Road-cut | Core POI, road segment | 0.290 | 0.351 | 4.3 |
| U-Net | Remote sensing images | 0.557 | 0.241 | 45.9 |
| UperNet | Remote sensing images | 0.532 | 0.347 | 65.3 |
| DETR | Remote sensing images | 0.561 | 0.234 | 75.5 |
| AOITR | Remote sensing images, core POI and spatial reference points | **0.726** | **0.497** | 62.0 |

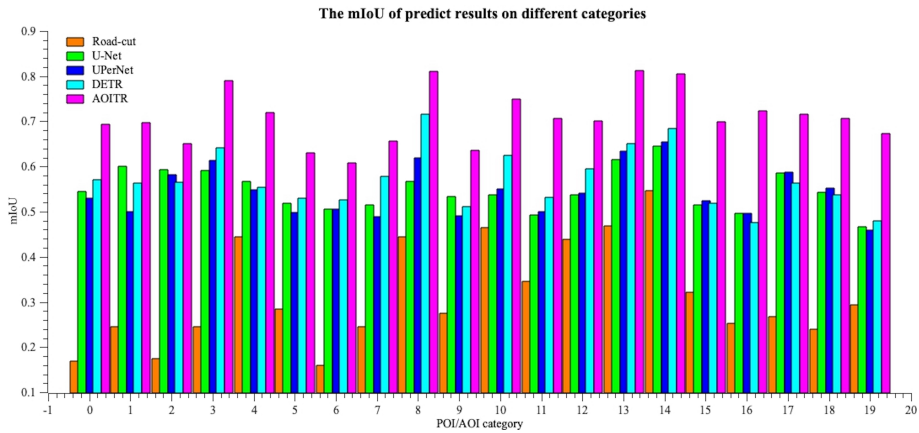

**Figure 6: The mIoU of predicted results on different Categories.**

Furthermore, we analyzed the mIoU metric of our AOITR and other baseline methods across different categories, as illustrated in Figure 6. Although the Road-cut algorithm performs well in well-developed areas regardless of the complexity required for multiple steps, it is ineffective in urban suburbs. On the other hand, semantic segmentation models based on pixel classification achieve good results on geographic base objects such as buildings or school playgrounds [11, 25], but their predicted boundaries in AOI detection are unusable for real-world businesses.

While our AOITR multimodal algorithm has a superior overall performance, the prediction accuracy varies significantly across different categories. For instance, Figure 7(a-c) depict residential areas, primary and secondary schools, and Industrial Parks with a mIoU exceeding 0.8. The ground truth is represented by the polygons outlined in red, while the detecting results are shown as green polygons. However, for factories, high-rise office buildings, hotels in dense areas, and large or complex AOIs that intersect external roads, the prediction accuracy requires further improvement. Figure 7(d-f) illustrates three failed examples of AOI generation.

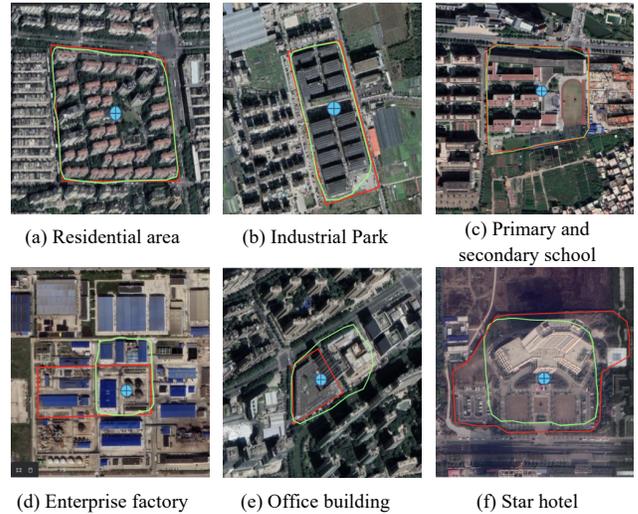

(a) Residential area  (b) Industrial Park  (c) Primary and secondary school

(d) Enterprise factory  (e) Office building  (f) Star hotel

**Figure 7: The predicted capabilities of AOITR in different categories.**

*4.3.2 Reliability evaluation of the predicted AOI.* The cascade module is a supervised binary classification model used to predict the reliability of the generated AOI polygon. The precision and recall metrics of the model are illustrated in the Figure 8 below:

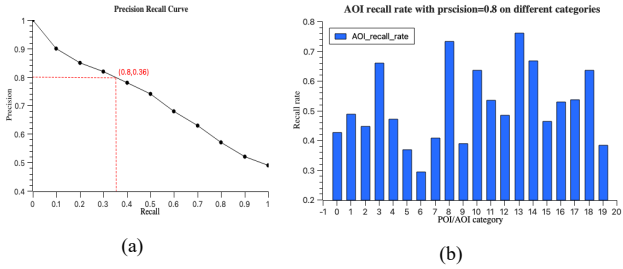

Figure 8: The precision and recall metrics of all categories.

Figure 8(a) presents the precision-recall curve for predicting AOI polygon of all categories generated from the AOITR. The curve indicates that approximately 49% of the AOIs we predicted are of high quality. However, for some demanding business scenarios, we require an accuracy rate of at least 80%. Hence, we will recall about 36% of the AOIs, as shown in the red dashed lines. Figure 8(b) illustrates the final recall rate of different categories. We observe that the AOI of category code 3, 8, 13, and 14 (i.e., Gas/other energy station, Industrial Park, Primary and secondary school, and Residential area) exhibits better detection capabilities. Conversely, category code 5, 6, 7, and 9 (i.e., Kindergarten, Star Hotels, Leisure and entertainment place, and Car service area, parking lot) display relatively poor results. The reasons for the poor predictive ability of some categories or specific cases can be summarized as follows:

- When the AOI comprises multiple semantic elements with complex boundaries, it increases the prediction difficulty, and the detection results may lose some subsidiary areas.
- When the core POI location is far from the actual AOI polygon center, it can affect the sampling of geographical prior. Especially in cases where the buildings are dense and the ground truth AOI area is relatively small.

### 4.4 Ablation analysis

In the AOITR, we incorporated a range of multimodal data to predict the polygon of the AOI for the input core POI. To assess the influence of different geographical prior information on AOI polygon generation, we conducted necessary ablation experiments, which are presented in Table 3. In the table, (w/o) denotes the condition "without".

Table 3: The multimodal ablation experiments.

| Modality | Total mIoU | Total high-IoU rate | Loss convergence epoch |
|---|---|---|---|
| (w/o) core POI location | 0.558 | 0.285 | 30 |
| (w/o) core POI category | 0.654 | 0.432 | 30 |
| (w/o) road reference points | 0.657 | 0.433 | 30 |
| (w/o) remote sensing images | 0.290 | 0.351 | - |

The ablation experiments show that the core POI location is the most important feature for predicting the AOI polygon. This is probably because the core POI location is typically situated at the center of the AOI polygon. The core POI category and road reference points have a similar impact, lowering the mIoU indicator by approximately 0.07. In the final experiment, where remote sensing images are abandoned, the Road-cut algorithm is employed to generate the AOI polygon, leading to the worst results.

Furthermore, we performed experiments to investigate the impact of different N values on the results. Specifically, we explored the outcomes when N was set to 4, 8, 16, and 32, simultaneously analyzing the impact of indicators mIoU and loss convergence speed. The results are depicted in Figure 9 below:

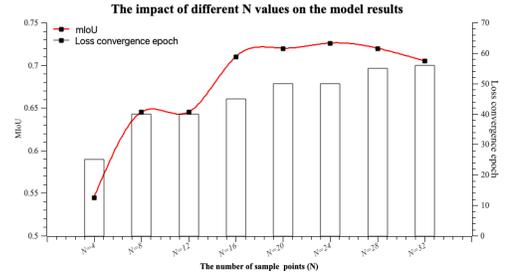

Figure 9: The effect of different N values on the results of mIoU.

Observing Figure 9 above, it is apparent that increasing N leads to better mIoU prediction results, albeit with diminishing returns and slower convergence speed during model training. However, it also suggests that N=24 could be the optimal choice.

## 5 Conclusions

This work is based on the demands from Ant Group's real-world O2O businesses. In order to timely and accurately generate AOI on a large scale, we propose an end-to-end multimodal AOI detection framework using remote sensing image and geographical prior. Our research aims to predict the AOI polygon given the core POI, and evaluate its reliability through the transformer-based AOI detecting model (AOITR). To validate our approach, we leverage both open-source and our internal POI, AOI data, along with other proprietary multimodal datasets and remote sensing images. We compare our model and other well-established baselines, including the famous Road-cut, semantic segmentation and object detection algorithms, on mIoU metric. Our proposed AOITR significantly outperforms the compared baselines. Finally, we have performed an ablation experiment to explore the effects of different geographical reference information on the performance of AOI detection. The AOIs produced by proposed AOITR have substantially enriched our AOI application library and have been successfully applied on over 10 different O2O businesses within Ant Group including Alipay's face scan payment service.

For future work, we believe the current progress of Large Language Models (LLMs) enables more relevant data sources, particularly textual data, to incorporate into our multimodal framework, further improving the overall AOI generation accuracy.